\documentclass[runningheads]{llncs}
\usepackage{graphicx}
\usepackage{caption} 
\usepackage{algorithm}
\usepackage{algpseudocode} 
\usepackage{amssymb,mathtools}

\begin{document}
\title{Binary Classification using Pairs of Minimum Spanning Trees or N-ary Trees}

\author{Riccardo La Grassa\inst{1}\orcidID{0000-0002-4355-0366} \and
Ignazio Gallo\inst{1}\orcidID{0000-0002-7076-8328} \and
Alessandro Calefati\inst{1}\orcidID{0000-0003-3860-4785} \and 
Dimitri Ognibene\inst{2}\orcidID{0000-0002-9454-680X}}
\authorrunning{R. La Grassa et al.}

\institute{University of Insubria, Varese, Italy \and
University of Essex, Colchester, Essex
}
\maketitle 
\begin{abstract}
 One-class classifiers are trained only with target class samples. Intuitively, their conservative modeling of the class description may benefit classical classification tasks where classes are difficult to separate due to overlapping and data imbalance. 
 In this work, three methods leveraging on the combination of one-class classifiers based on non-parametric models, N-ary Trees and Minimum Spanning Trees class descriptors (MST\_CD) are proposed.

 These methods deal with inconsistencies arising from combining multiple classifiers and with spurious connections that MST-CD creates in multi-modal class distributions. 
 Experiments on several datasets show that the proposed approach obtains comparable and, in some cases, state-of-the-art results.

\keywords{One-class Classifiers \and Minimum Spanning Tree \and Instance-based approaches \and Non-Parametric models}
\end{abstract}

\section{Introduction}
With the rise of social platforms and internet, data produced by users has grown exponentially enabling the use of data greedy machine learning algorithms in several applications, still in many domains of practical interests, data are still scarce and require more data efficient methods especially for non-trivial classification tasks where classes are difficult to separate due to overlapping and data imbalance.
One-class classifiers are trained with target class only samples under the strong assumption that data from the other classes are not available or have low quality. 
Intuitively, their conservative modeling of the data distribution may benefit classical classification tasks if they were combined in a effective manner.
 
In this work, three methods are proposed which leverage on the combination of one-class classifiers based on non-parametric models, K-Nearest Neighbour,  Trees and Minimum Spanning Trees class descriptors (MST-CD), to tackle binary classification problems. 
 

In the first model, we train classifiers using Minimum Spanning Tree Class Descriptor (MST\_CD) in the training step and then apply a new technique to provide a more reliable prediction. 
The second model creates a more powerful classifier based on MST\_CD combining results according to an ensemble method. 
The third model is very similar to the previous one but uses a  tree starting to the closest neighbour to the target pattern for each classifier and finally it leverages on the ensemble technique. 

In the next Section related works are shown, the proposed approach is described in Section~\ref{sec:proposed-approach}, Experiments are in Section~\ref{sec:experiments} and Conclusions in~\ref{sec:conclusions}.

\section{Related work}\label{sec:related_works}
In the feature selection field, many approaches have been proposed and used with classifiers to obtain better accuracy~\cite{Budak,2018:Krakovna:interpretable,2019:Abpeykar,Singh}.
Krawczyk~\textit{et al.}~\cite{Krawczyk} proposes a generic model that improves the performance of many common classifiers showing des standard and des-threshold methodologies. 
They are based on a k-Nearest Neighbour (k-NN) technique that assigns a pattern to a class on the basis of the class of its nearest $k$ neighbours.
The approach shows good results and authors compare it with classical models. 
However, they do not consider the case in which no classifier is “activated” and, thus, when it is not possible to obtain a prediction for a new instance $x$.
Duin~\textit{et al.}~\cite{tax2002using} proposes a simple approach to assign these refused objects to the class with largest prior probability, but they do not describe a method in the scenario where the two decision boundaries overlap.
Our approach combines part of the approach described in~\cite{Krawczyk}\cite{tax2002using} using MSTs with other methodologies to improve the accuracy and considering also the overlapping.
Abpeykar~\textit{et al.}~\cite{2019:Abpeykar} proposes a survey that sums up the performance of many classifiers on well-known datasets from UCI repository.
A milestone on one-class classifier comes from Pekalska~\textit{et al.}~\cite{Pekalska} with their MST descriptor. 
The original idea was to try to search a pattern from all training sets in order to create a MST that represents the model on which will be done some geometrical operations with the goal to generate a border for a specific class.
Segui~\textit{et al.}~\cite{Segui} focuses on the research of noise within a target class and removes it in order to have better accuracy at testing time. 
They confirm that a graph-based one-class classifier, like MST\_CD obtains good results than other approaches, especially dealing with small samples cardinalities and high data dimensionalities. 
Quinlan~\cite{Quinlan} proposes a general method that allows predictions using both mixed approach of instance-based and model-based learning. 
He proves that these composite methods often produce better results in term of predictions than using only a single methodology.

\section{Minimum Spanning Tree class descriptor}
As widely described in~\cite{Pekalska} a MST\_CD is a non-parametric classifier able to create a structure, seeing only data of the single class of interest. 
This structure is based on Minimum Spanning Tree, basic elements of this classifier are not only vertices but also edges of the graph, giving a richer representation of the data. 
Considering edges of the graph as target class objects, additional virtual target objects are generated. 
The classification of a new object $x$ is based on the distance to the nearest vertex or edge.
The key of this classifier is to track a shape around the training set not considering only all instances of the training but also edges of the graph, in order to have more structured information. 
Therefore, in prediction phase, the classifier considers two important elements:

\begin{itemize}
    \item Projection of point $x$ on a line defined by vertices ${x_i, x_j}$
    \item Minimum Euclidean distance between $(x, x_i)$ and $(x, x_j)$
\end{itemize}

The Projection of $x$ is defined as follow:
$$p_{e_{i,j}}(x)=x_i + \frac{(x_j - x_i)^T(x_j - x_i)}{||x_j - x_i||^2}(x_j - x_i)$$
if $p_{e_{i,j}}(x)$ lies on the edge $e_{i,j}$= ($x_i$,$x_j$), we compute $p_{e_{i,j}}(x)$ and the Euclidean distance between $x$ and $p_{e_{i,j}}(x)$, more formally:
$$0<=\frac{(x_j - x_i)^T(x_j - x_i)}{||x_j - x_i||^2}<=1$$ then
$$d(x|e_{i,j})=||x - p_{e_{i,j}}(x)||$$
Otherwise we compute the Euclidean distance of $x$ and pairs ($x_i$, $x_j$), precisely:
$$d(x|e_{i,j})=min(||x - x_j||,||x - x_i||)$$

Therefore, a new object $x$ is recognized by MST\_CD if it lies in proximity of the shape built in training phase, otherwise the object is considered as outlier.
The decision whether an object is recognized by classifier or not is based on threshold of the shape created during the training phase, more formally:
$$d_{MST\_CD}(x|X) <= \theta$$
Authors set the threshold $\theta$ as the median of the distribution of the edge weights $w_{ij}=||e_{ij}||$ in the given MST. 
Given $\hat{e}=(||e_1||,||e_2||,...,||e_n||)$ as an ordered edge weights values, they define $\theta$ as $\theta = ||e_{[\alpha n]}||$, where $\alpha \in [0,1]$. 
For instance, with $\alpha=0.5$, we assign the median value of all edge weights into the MST.

\section{Proposed Approach}\label{sec:proposed-approach}
The main objective of a one-class classifier is to recognize instances of a selected class from a set of samples.
All instances that are not classified by this model will be considered as outliers (or alien class), while others will be recognized as belonging to the same class of the training set. 
In this context, we cannot say anything about the refused objects (outlier), but if we have a one-class classifier for each label of the dataset we can not have outliers because if a classifier refuses an object, it should be accepted by the others classifiers.
In this work, we use two one-class classifiers trained on two different classes.
We assume a discriminant function $f_{ab}$ on a binary classification problem to classify a new object considering acceptation and rejection from both our classifiers. 
Given $X$ a MST and $x$ a new object, we define a function that assigns a label such that:
\begin{equation}
  f_{ab}(x,X) =
  \begin{cases}
    1 & \text{if  $d_{MST\_CD_0}(x|X) <= \theta$ and $d_{MST\_CD_1}(x|X) > \theta_1$} \\
    0 & \text{if  $d_{MST\_CD_0}(x|X) > \theta$ and $d_{MST\_CD_1}(x|X) <= \theta_1$}
  \end{cases}
\end{equation}
However, we may have two anomaly cases:
\begin{enumerate}
    \item both classifiers refuse the pattern to be predicted
    \item both classifiers accept the pattern to be predicted
\end{enumerate}
In this specific case we can apply a simple technique to assign a pattern to two possible classes depending on its distance to clusters. 
More formally, given $x$ a target vector, $i$ and $j$ are all elements of the dataset belonging to both classes, we define:

$$d(x|i)=(|x-i_{0}|,|x-i_{1}|,...,|x-i_{n_{0}}|)$$
$$d'(x|j)=(|x-j_{0}|,|x-j_{1}|,...,|x-j_{n_{1}}|)$$

where $n_0$ and $n_1$ are cardinalities of first and second class respectively.

Then we take the $k$ nearest elements for each class, such that:

$$k <= min(n_0, n_1)$$

We compute the vector difference as:

$$Vector_{diff}=[(d_1-d'_1),(d_2-d'_2),...,(d_k-d'_k)]$$

Finally, We assume a new function $f'_{ab}$ to classify the
object as:
\begin{equation}
  f'_{ab}(x,Vector_{diff}) =
  \begin{cases}
    1 & \text{if } |Vector_{diff} (k<=0)| > |Vector_{diff} (k>0)| \\
    0 & \text{otherwise}
  \end{cases}
\end{equation}


\subsection{First approach}

Our first approach combines two one-class classifiers based on Minimum Spanning Tree class descriptor and solve both above mentioned issues.
When one of the two issues appear, e.g. both classifiers accept (classifiers overlap) or reject (uncovered) the input sample, an approach similar to K-NN majority vote is applied.
Using the already computed euclidean distances  between the sample and the elements of the two MSTs, the K elements  of each MST ($MST_1$,$MST_2$) nearer to the sample are selected to check which of the two MSTs is consistently closer to the sample. This is done by internally sorting the two sets of K elements in increasing distance order from the sample then subtracting the 2 corresponding K-ary vectors ($D_1$, $D_2$) of distances and finally counting how many positive elements are in the resulting vector $R=D_1-D_2$. If $R$ contains more positive than negative elements the sample is associated to $MST_2$ as its elements are closer to the sample. This method allows to integrate the generalization capabilities of the MST with the robustness offered by the K-NN vote strategy to deal with binary classification strategies. See the pseudo-code in Algorithm~\ref{alg:MST_CD_OVA}.

\begin{figure}[hbt]
\centering
\includegraphics[scale=0.35]{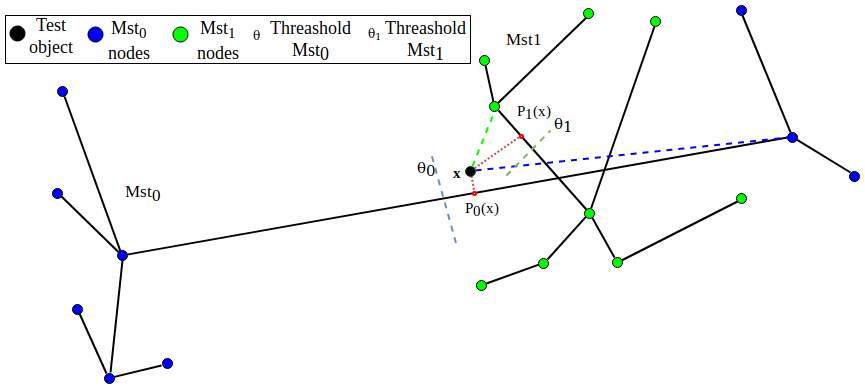}
\caption{MST\_CD model: In this figure, we show the problem about the competence area into two different classifiers in a toy scenario. 
In this case the object $x$ will be predicted as blue class instead of green class. 
This happens because the orthogonal projection of $x$ is nearer to an edge that contains an outlier and the object is located into a non-competence area of the blue-classifier making a wrong prediction.}\label{fig:fig1}
\end{figure}
\begin{algorithm} \caption{MST CD OVA}
\label{alg:MST_CD_OVA}
\scriptsize
\begin{algorithmic}[1]
\For {$v \in \mathcal G_0 $}
\State all euclidean distances $\gets || x - v ||$
\EndFor
\State NodeX = Take min(all euclidean distances)
\State EdgeNodeX $\gets$ Search inc/out edge nodeX and return $(u,v)$
\For {$u,v \in \mathcal E(NodeX)$ }
\If {$0<=\frac{(x_j - x_i)^T * (x-x_i)}{||x_j - x_i||^2}<=1$}
\State {$P_{e_{_{i_{j}}}}(x)=x_{i} + \frac{(x_{j} - x_{i})^T * (x-x_i)}{||x_j - x_i||^2}*(x_j-x_i)$}
\State {$d(x|e_{_{i_{j}}}) \gets ||x - P_{e_{_{i_{j}}}}(x)||$}
\Else
\State{$d(x|e_{_{i_{j}}}) \gets min \big\{ ||x - x_i||, ||x -x_j||\big\}$}
\EndIf
\EndFor
\State \textbf{Repeat line 1-13 for graph $G_1$}
\State min dist0 = $min(d(x|e_{ij}))$
\State min dist1 = $min(d1(x|e_{ij}))$
\State 1 $\gets$  \text{if  $d_{MST\_CD_0}(x|X) <= \theta$ and $d_{MST\_CD_1}(x|X) > \theta_1$} \\
0 $\gets$ \text{if  $d_{MST\_CD_0}(x|X) > \theta$ and $d_{MST\_CD_1}(x|X) <= \theta_1$}
\If{\textbf{min dist0 $<= \theta$ and min dist1 $<= \theta_1$}} 
\State knn weight1=order($d_{1}(x|u)$) and take $k_1$-elements 
\State knn weight0=order($d_{0}(x|v)$) and take $k_1$-elements
\State euclidean distance vectors = (knn weight1 - knn weight0)
\State positive $=$ Count $n_i > 0$ in euclidean distance vectors
\State negative $=$ Count $n_i < 0$ in euclidean distance vectors
\If{negative $>=$ positive}
\State prediction $\gets 1$
\Else
\State prediction $\gets 0$
\EndIf
\EndIf
\If{\textbf{min dist0 $> \theta$ and min dist1 $> \theta_1$}}
\State The approach is equal to lines [20-29]
\EndIf
\end{algorithmic}
\end{algorithm}

\subsection{Second approach}
\begin{figure}
\centering
\includegraphics[scale=0.3]{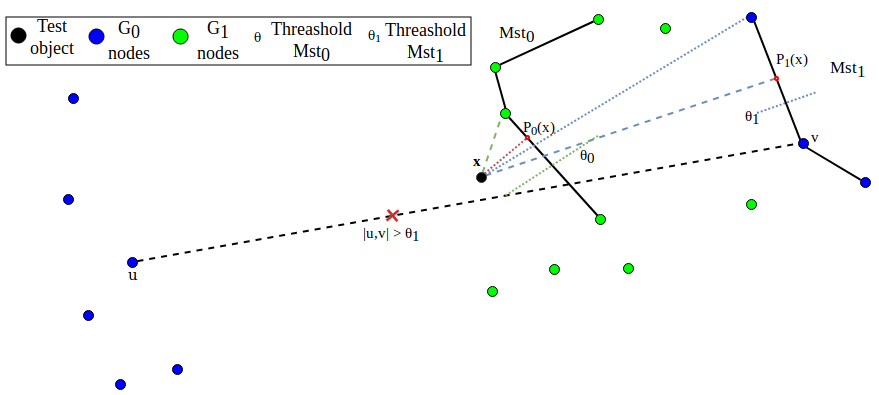}
\caption{MST\_CD\_GP model: A different approach on the same toy example considering a combination built on two small MST starting from nodes with a minimum distance from target pattern.} \label{fig:fig2}
\end{figure}

In a second approach, we mix K-NN and MST in the reverse order. Using a K-NN like approach, we initially select the K  training samples from each class closest to the X sample to classify. This results in two sets $S_1$ and $S_2$, we then create two K-elements MST for each of this sets ($kMST_1$,$kMST_1$). 
After this we classify X as we did for approach 1. This second approach has two advantages over the previous. One, is that it does not need to perform the expensive creation of the two large MST covering the whole dataset (quadratic in the number of samples) but only deals with a K-elements MSTs. Second, it avoids MST spurious  edges between elements of  distant and unrelated  modes of a class distribution. See the pseudo-code in Algorithm~\ref{alg:MST_CD_GAMMA} and Fig.\ref{fig:fig2}.

\begin{algorithm} \scriptsize
\caption{MST CD with Gamma parameters}
\begin{algorithmic}[1]
\Function{create small MST}{g0 weight sorted, g1 weight sorted}
\State $small G_0$ = first gamma index sorted values in g0 weight sorted
\State edges couple $\gets$ all combinations nodes small g0
\For {$u,v \in \mathcal edges couple$}
\State $small\ G_0 \gets (node\ u, node\ v, weight=(u,v))$
\EndFor
\State $small\ MST_0 = ComputeMST(small\ G_0)$
\State Same approach for $MST_1$
\State $e(small\ MST 0) = (||e_0||,||e_1||,..||e_n||)$
\State $e(small\ MST 1) = (||e_0||,||e_1||,..||e_n||)$
\State $\theta_0= || e_{(\alpha n )}||$
\State $\theta_1= || e_{(\alpha n )}||$\\
\Return small $MST_0$, $small MST_1$
\EndFunction
\end{algorithmic}
\label{alg:MST_CD_GAMMA}
\end{algorithm}

\subsection{Third approach}
\begin{figure}[hbt]
\centering
\includegraphics[scale=0.35]{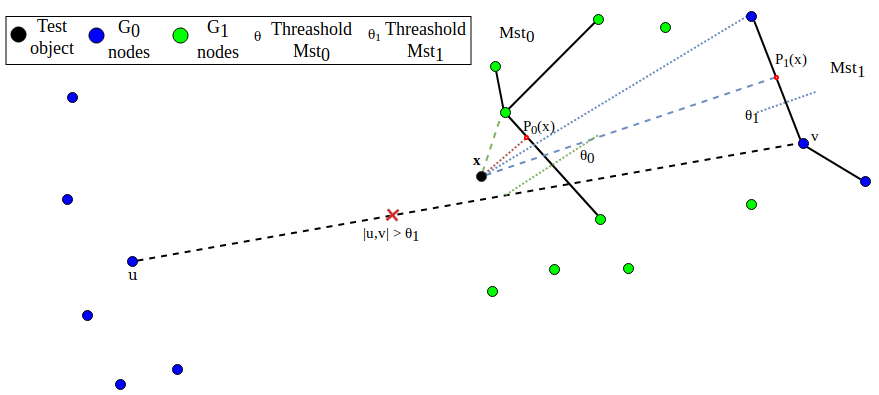}
\caption{N-ary model: A different approach on the same toy example.
Using a N-ary model makes less complex the prediction phase ignoring outliers from the structure.}\label{fig:fig3}
\end{figure}

The third is similar to the second one but instead of selecting the $K$ nearest elements to the sample $X$ to classify, it selects only the nearest element $E$ to the sample and then selects the K-1 elements of the training set nearest to $E$. 
It then creates a tree having $E$ as root and it uses it as a classifier like in previous cases. 
This further extends the robustness to outliers and spurious connections between far nodes.
See the pseudo-code in Algorithm~\ref{alg:MST_CD_N_ARY} and Fig.\ref{fig:fig3}.
\begin{algorithm}
\scriptsize
\caption{N-ary model}
\label{alg:MST_CD_N_ARY}
\begin{algorithmic}[1]
\Function{create\_n\_ary}{$sub G_0 \ weight\ sorted, sub G_1\  weight\ sorted, \ gamma\ index$}
\State $n\_ary_0 = first\ gamma \ index \ sorted \ values \ in \ g0 \ weight\  sorted$
\State $n\_ary_1 = first\ gamma \ index \ sorted \ values \ in \ g1 \ weight\  sorted$
\For {$u,v \in \mathcal sub G\_0 weighted sorted $}
\State $n\_ary_0 \gets (node u, node v, weight=(u,v))$
\EndFor

\For {$u,v \in \mathcal sub G\_1 weighted sorted$}
\State $n\_ary_1 \gets (node u, node v, weight=(u,v))$
\EndFor

\State \textbf{return} $n\_ary_0, n\_ary_1$
\EndFunction
\end{algorithmic}
\end{algorithm}

\section{Datasets}
\label{sec:datasets}
We use five low-dimensional datasets (\textit{Hill}, \textit{Sonar}, \textit{Australian}, \textit{Mofn}, \textit{Pima}) and three high-dimensional datasets (\textit{Arcene}, \textit{Gisette}, \textit{Madelon}), all taken from the UCI repository~\cite{Dua:2019}. 
All of them have two classes.
Table~\ref{tab:dataset} shows the details of all datasets.
To evaluate our approach in a more robust way, we selected datasets with a huge variability in term of number of features and number of instances.
For all the datasets we use 5-fold Cross Validation. 
We replace the missing values in the datasets with the average value for the missing features.

\begin{table}
\centering
\caption{Number of features, classes, instances and positive-negative samples for all the datasets used in our experiment.}\label{tab:dataset}
\begin{tabular}{l|cccc} 
\hline
Datasets         & Features & Classes & Instances & pos-neg  \\ \hline
Arcene           & 10000    & 2       & 100     & 44-56  \\ \hline
Gisette          & 5000     & 2       & 6000    & 3000-3000  \\ \hline
Madelon          & 500      & 2       & 2000    & 1000-1000 \\ \hline
Hill             & 101      & 2       & 606     & 305-301 \\ \hline
Sonar            & 60       & 2       & 208     & 97-111 \\ \hline
Australian       & 14       & 2       & 690     & 307-383 \\ \hline
Mofn             & 10       & 2       & 1324    & 292-1032 \\ \hline
Pima             & 8        & 2       & 688     & 305-301 \\ \hline
\end{tabular}
\end{table}

\section{Experiments}\label{sec:experiments}
In this section we report two groups of experiments, 
\begin{itemize}
    \item to study the effect on the parameters in our models,
    \item to compare the proposed models with the results available in the literature on some well known benchmarks.
\end{itemize}
\subsection{Parameters evaluation}
For each classifier, we extract some common metrics to evaluate the performance, such as Sensitivity, Precision and F1 Score. 
Starting from these values we create confusion matrix and obtain the final accuracy considering True/False positive and True/False negative samples. 
\begin{table}[htb]
\centering
\caption{These are the results of both classifiers using our MST\_CD model about Sensitivity, Precision, F1 Score and Total accuracy on Sonar dataset. Last line contains the average for each column.}
\begin{tabular}{cccccccc}
\hline
\multicolumn{8}{c}{\textbf{MST\_CD}}                                                                               \\
$Fold$ & $Sensitivity^0$ & $Sensitivity^1$ & $Precision^0$ & $Precision^1$ & $F1^0$ & $F1^1$ & $Accuracy$  \\\hline
1       & 0,867         & 0,900         & 0,867       & 0,900       & 0,867      & 0,900      & 0,886     \\
\hline
2       & 0,833         & 0,944         & 0,938       & 0,850       & 0,882      & 0,895      & 0,889     \\
\hline
3       & 0,800         & 1,000         & 1,000       & 0,833       & 0,889      & 0,909      & 0,900     \\
\hline
4       & 0,688         & 0,880         & 0,786       & 0,815       & 0,733      & 0,846      & 0,805     \\
\hline
5       & 0,714         & 0,850         & 0,769       & 0,810       & 0,741      & 0,829      & 0,794     \\
\hline
\textit{Average} & 0,780         & 0,915         & 0,872       & 0,842       & 0,822      & 0,876      & \textbf{0,855}    \\\hline
\end{tabular}
\end{table}
\begin{table}[htb]
\centering
\caption{Details of our third model N-ary considering the same measures than above table. We show an accuracy of 87.3\% with gamma=20 and gamma=10 in N-ary model against 85.5\% in our first model. Results of each lines are extracted using an average on k=5 fold-validation.}
\label{tab:n_ary_gamma}
\centering
\begin{tabular}{cccccccc}
\hline
\multicolumn{8}{c}{\textbf{N-ary}}                                                                               \\
$Gamma$ & $Sensitivity^0$ & $Sensitivity^1$ & $Precision^0$ & $Precision^1$ & $F1^0$ & $F1^1$ & $Accuracy$  \\
\hline
30    & 0,791         & 0,915         & 0,873       & 0,850       & 0,829      & 0,881      & 0,860     \\
\hline
20    & 0,793         & 0,915         & 0,873       & 0,850       & 0,830      & 0,881      & 0,861     \\
\hline
10    & 0,820         & 0,915         & 0,879       & 0,869       & 0,847      & 0,890      & \textbf{0,873}     \\
\hline
8     & 0,830         & 0,906         & 0,869       & 0,876       & 0,847      & 0,890      & \textbf{0,873}     \\
\hline
6     & 0,816         & 0,880         & 0,849       & 0,854       & 0,830      & 0,865      & 0,851     \\
\hline
4     & 0,808         & 0,882         & 0,852       & 0,848       & 0,826      & 0,863      & 0,848     \\
\hline
3     & 0,787         & 0,861         & 0,830       & 0,832       & 0,804      & 0,844      & 0,829    \\\hline
\end{tabular}
\end{table}
\begin{table}[htb]
\centering
\caption{Parameters set}
\label{tab:parameters set}
\begin{tabular}{ccccccccc} 
\hline
Parameters & Arcene & Madelon & Gisette & Mofn & Australian & Pima & Sonar & Hills  \\ 
\hline
Threshold  & 0.25   & 0.25    & 0.2     & 0.05 & 0.05       & 0.05 & 0.1   & 0.9    \\ 
\hline
Neighbours & 4      & 12      & 4       & 2    & 6          & 6    & 2     & 2      \\
\hline
\end{tabular}
\end{table}
\begin{figure}[htb]
\centering
\includegraphics[width=0.6\textwidth]{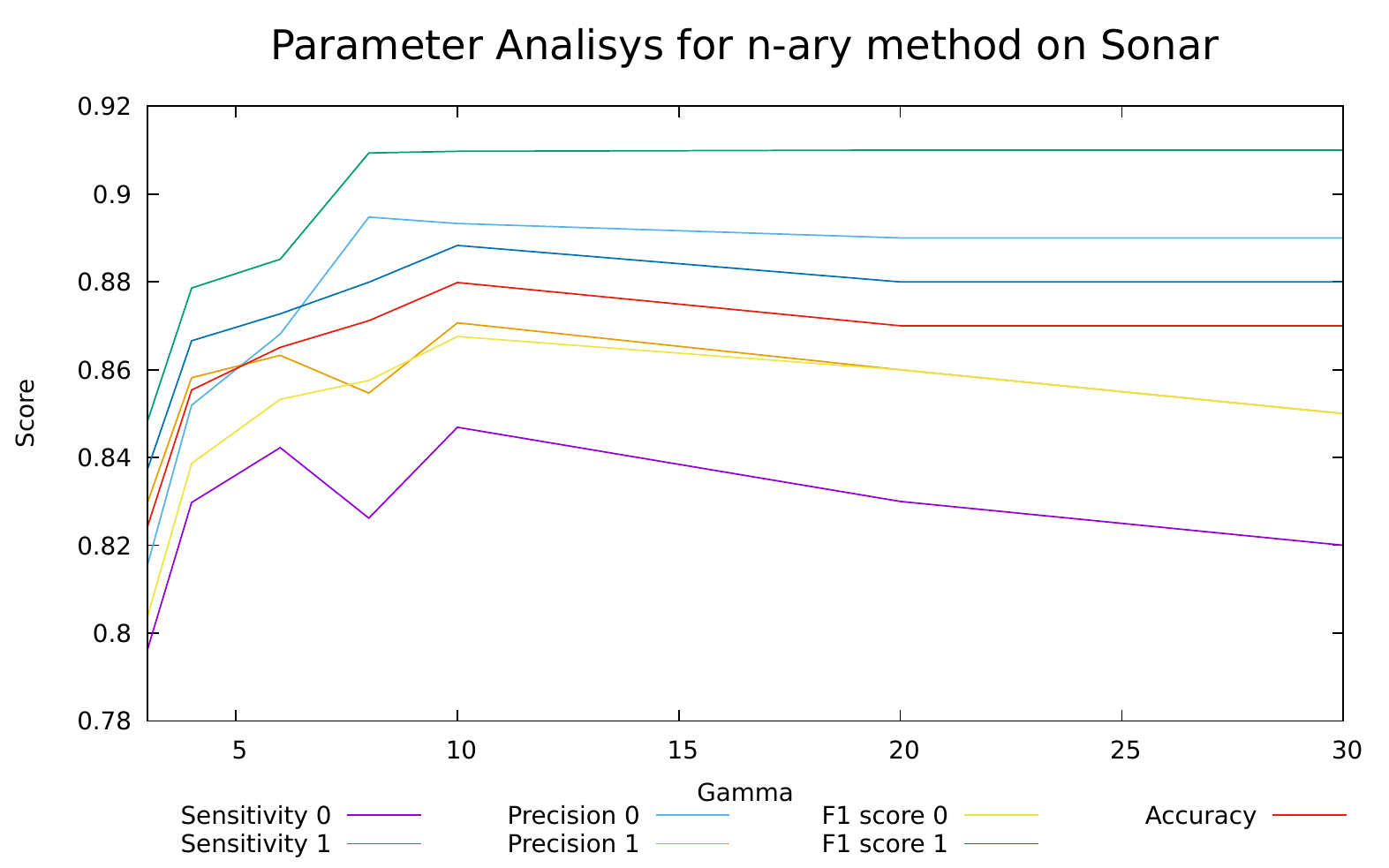}
\caption{Best Gamma parameter search for  method using the Sonar dataset.
We use the average accuracy to select the best parameter.} \label{fig:parameter_analisys_n-ary}
\end{figure}
\subsection{Comparison}
We evaluated the proposed models on the well known benchmark datasets described in Section~\ref{sec:datasets}.
In particular, we compared our methods using the results published in~\cite{2018:Krakovna:interpretable,2019:Abpeykar} and shown the comparison in Tables~\ref{tab:cv_comparison_with_krakovna},~\ref{tab:cv_comparison_ensemble} and~\ref{tab:cv_comparison_ml}. 
In our work, we have two one-class classifiers and the rate of True/Negative samples has been computed considering two different models. 
For instance if an object has been predicted correctly True positive by a classifier the others must be refuse and label it as True negative, otherwise, we consider all the possible combinations ($2^{2}$) in our case (two classifiers). 
The final accuracy is obtained from average of Cross-Validation.
In our experiments, we set $\gamma$ parameter into the second and third model within the range [2,30] nodes and take the maximum accuracy obtained from Cross-Validation methods considering always the same dimension training/test (80\%-20\%) see Table \ref{tab:n_ary_gamma}. 
Threshold measures and k-neighbours are set as we show into Table \ref{tab:parameters set} on each specific dataset. 
In Table~\ref{tab:cv_comparison_with_krakovna} we compare results of our models with many references used to examine the performance of classifiers (Moradi \& Rostami, 2015).
Our models do not use feature selection techniques, therefore for each instance we exploit all available features. 
Comparing our models with the ensemble methods known as AdaboostM1~\cite{2002:Eibl}, Bagging~\cite{1997:Ting}, Dagging~\cite{1997:Ting}, LogitBoost~\cite{2002:Frank}, MODLEM~\cite{1998:Stefanowski}, Decorate (Melville \& Mooney, 2003), Grading~\cite{seewald2001evaluation}, MultiBoostAB~\cite{webb2000multiboosting} and StackingC (Seewald, 2002), our models overcome the accuracy of other ensembles classifier (except for Arcene dataset where Bagging, LogitBoost and Modlem have higher accuracy than our models).

\begin{table}
\centering
\caption{5-fold Cross-Validation accuracy results on $6$ datasets. 
Results obtained with our three models (MST\_CD,  MST\_CD\_GP, N-ary) are compared with results published in~\cite{2018:Krakovna:interpretable} on same datasets.}

\label{tab:cv_comparison_with_krakovna}
\resizebox{\columnwidth}{!}{
\begin{tabular}{c|cccccccccc|ccc} 
\hline
 & \multicolumn{10}{|c|} {Krakovna \textit{et al.}~\cite{2018:Krakovna:interpretable}}  & \multicolumn{3}{c}{Our models} \\ \hline
Dataset                & Bart & c5.0 & Cart & Lasso & LR   & NB   & RF   & SBFC & SVM  & TAN  &  CD     & CD\_GP  &   N-ary  \\ \hline
Arcene                 & 71.6 & 66   & 63   & 65.6  & 52   & 69   & 71.8 & 72.2 & 72   & -    & 79.6  & 77.7   & \textbf{80.3}    \\ \hline
Madelon                & 76   & 75.8 & \textbf{ 78.2 }& 60.7  & 60   & 59.8 & 67.1 & 63.4 & 62   & 54.2 & 75    & 75.2   & 75.3  \\ \hline
Gisette                & \textbf{97.7} & 94.8 & 90.8 & 97.2  & 88.1 & 90.3 & 97   & 95.2 & 96.9 & -    & 96.8  & -      & -  \\ \hline
Mofn                   & \textbf{100}  & 84.8 & 83.9 & \textbf{100} & \textbf{100}  & 86.4 & 92.4 & 86.2 & 94.6 & 92.1 & \textbf{100}   & \textbf{100} & \textbf{100} \\ \hline
Australian             & 86.9 & 86.7 & 84.2 & 85.6  & 86.8 & 85.7 & \textbf{87.8} & 86.9 & 86   & 86.8 & 69.0  & 70.6   & 70.3 \\ \hline
Pima                   & 78.2 & 76.8 & 75.7 & 78    & 78.3 & 78   & 77.8 & \textbf{78.9} & 78.1 & \textbf{78.9} & 70.1  & 71.7   & 74.1 \\ \hline
\end{tabular}
}
\end{table}

\begin{table}
\centering
\caption{5-fold Cross-Validation accuracy results on five different datasets. 
Results obtained with our three models (MST\_CD,  MST\_CD\_GP, N-ary) are compared with the results of different ensemble methods published in~\cite{2019:Abpeykar}.}\label{tab:cv_comparison_ensemble}
\resizebox{\columnwidth}{!}{
\begin{tabular}{c|ccccccccc|ccc} 
\hline
  & \multicolumn{9}{|c|} {Abpeykar \textit{et al.}~\cite{2019:Abpeykar}}      & \multicolumn{3}{c}{Our models}         \\ \hline
Dataset  & AdaB. & Bagg. & Dagg. & LogitB. & Mod. & Decor. & Grad. & Mt.B & Stack.C & CD & CD\_GP & N-ary  \\ \hline
Arcene   & 79.5  & 82.5  & 74.5  & 85.5  & \textbf{86.0}   & -     & 56.0  & 80.0   & 56.0  & 79.6   & 77.7  & 80.3   \\ \hline
Madelon  & 63.4  & 75.0  & 57.2  & 63.0  & 52.5   & 73.3  & 50.1  & 61.7   & 50.1  & 75     & 75.2  & \textbf{75.3}   \\\hline
Sonar    & 71.6  & 76.9  & 69.7  & 79.3  & 70.6   & 84.1  & 53.3  & 74.5   & 53.3  & 85.4   & 85.2  & \textbf{88}     \\ \hline
Hills      & 50.4  & 50.2  & 50.4  & 50.4  & -      & -     & 50.4  & 50.4   & 50.4  & 58.1   & 57.9  & \textbf{61.1}   \\ \hline
Gisette  & 88.9  & 75.0  & 82.2  & 89.4  & -      & 82.2  & 48.1  & 82.7   & 48.1  & \textbf{96.8}   & -     & -      \\ \hline
\end{tabular}
}
\end{table}

\begin{table}
\centering
\caption{5-fold Cross-Validation accuracy results on five datasets. 
The results obtained with our three models are compared with the results of different machine learning methods published in~\cite{2019:Abpeykar}.}\label{tab:cv_comparison_ml}
\begin{tabular}{c|ccccc|ccc} 
\hline
 & \multicolumn{5}{|c|}{Abpeykar \textit{et al.}~\cite{2019:Abpeykar}}  & \multicolumn{3}{c}{Our models}    \\ \hline
Datasets               & Dt & k-NN  & NB   & RF  & SVM  & MST\_CD & MST\_CD\_GP & N-ary   \\ \hline
Arcene                 & 67.2      & 66.7 & 69.7 & 71.4        & 71.4    & 79.6  & 77.7  & \textbf{80.3}  \\ \hline
Madelon                & \textbf{82.7}      & 74.8 & 79.3 & 77.9        & 78.4    & 75    & 75.2        & 75.3        \\ \hline
Sonar                  & 79.5      & 80.4 & 77.6 & 81.4        & 82.3    & 85.4  & 85.2  & \textbf{88.0}    \\ \hline
\end{tabular}
\end{table}

\section{Conclusion}\label{sec:conclusions}
The presented results show that the solutions proposed are competitive both with ensemble and classical classifiers. The second and third method that aimed at  increasing the robustness to outliers and spurious MST connections were proved to consistently ameliorate accuracy. Also the latter methods avoid the initial  computation of expensive large MST while requiring the computation of small MST at runtime and provide higher accuracy. This makes the latter methods more convenient with large datasets for which the complete MST computation may be too expensive. We are also confident that these methods can be highly optimized using caching methods to further improve online computation performance. Future work will tackle the issue of feature scaling and selection as well as the possibility to combine the approach with ensemble methods.

\bibliographystyle{splncs04}
\bibliography{egbib}
\end{document}